\documentclass[runningheads]{llncs}

\usepackage[T1]{fontenc}

\usepackage{siunitx}
\usepackage{amsmath}
\usepackage{graphicx}
\usepackage{multirow}
\usepackage{amsmath,amssymb,amsfonts}
\usepackage{mathrsfs}
\usepackage[title]{appendix}
\usepackage{xcolor}
\usepackage{colortbl}
\usepackage{textcomp}
\usepackage{manyfoot}
\usepackage{booktabs}
\usepackage{algorithm}
\usepackage{algorithmicx}
\usepackage{algpseudocode}
\usepackage{listings}
\usepackage{graphicx}
\definecolor{mistyrose}{rgb}{1.0, 0.89, 0.88}

\DeclareMathOperator*{\argmin}{arg\,min}

\begin{document}
\title{Data-Efficient Limited-Angle CT Using Deep Priors and Regularization}
\titlerunning{Limited-Angle CT Using Deep Priors and Regularization}
\author{Ilmari Vahteristo\inst{1} \and
Zhi-Song Liu\inst{1} \and
Andreas Rupp\inst{2}}
\authorrunning{Vahteristo, I. et al.}
\institute{LUT University, Lappeenranta 53850, Finland \and Saarland University,  66123 Saarbrücken, Germany}
\maketitle              % typeset the header of the contribution
\begin{abstract}
Reconstructing an image from its Radon transform is
a fundamental computed tomography (CT) task arising
in applications such as X-ray scans. In many
practical scenarios, a full 180-degree scan is not
feasible, or there is a desire to reduce radiation
exposure. In these limited-angle settings, the problem
becomes ill-posed, and methods designed for full-view
data often leave significant artifacts.
We propose a very low-data approach to reconstruct the
original image from its Radon transform under severe
angle limitations. Because the inverse problem is ill-posed,
we combine multiple regularization methods, including
Total Variation, a sinogram filter, Deep Image Prior,
and a patch-level autoencoder. We use a differentiable
implementation of the Radon transform, which allows
us to use gradient-based techniques to solve the inverse problem.
Our method is evaluated on a dataset from the Helsinki Tomography Challenge 2022,
where the goal is to reconstruct a binary disk from its
limited-angle sinogram. We only use a total of 12 data points--eight for learning a prior and four for hyperparameter selection--and achieve results comparable to the best synthetic data-driven approaches.

\end{abstract}
\section{Introduction}

% Background
Computed tomography (CT) is a widely used imaging technique
in medical diagnostics, industrial inspection, and scientific
research. It involves measuring the permeability of an object from multiple angles and reconstructing the object's interior from these projections. The Radon transform is a mathematical tool that describes how an object's structure
is projected onto a set of lines, forming the basis of CT
reconstruction \cite{beylkin1987discrete}.
In practice, the Radon transform is approximated by taking a finite number of projections at discrete angles, forming a sinogram.

% Motivation
Reconstructing an object from limited-angle data is a challenging problem due to the restricted range of projection angles. This situation arises in many practical scenarios,
such as industrial inspection where the imaging system cannot rotate
around the object a full 180$^\circ$, or in medical imaging where
patient positioning might limit scanner access. Because few angles
are captured, the problem is inherently ill-posed~\cite{hadamard2014lectures},
leading to multiple plausible images that match the measured data.

This work focuses on the Helsinki Tomography Challenge 2022 (HTC'22) \cite{meaney2023helsinki} which presents a difficult problem: reconstructing binary objects from limited angle data.
% According to Hadamard \cite{hadamard2014lectures}, an inverse problem is considered 'ill-posed' if no object exists that could produce the measurements, or if multiple objects could produce the same measurements, or if the solution does not change continuously with small changes in the data.
We introduce a gradient-based optimization pipeline that integrates multiple regularization techniques. We demonstrate that this method yields competitive reconstruction results using only 8 data points to train a patch-based regularizer--a substantial reduction compared to typical machine learning approaches. To promote reproducibility and future research, we provide an open-source implementation of our pipeline.

\section{Related works}
Reconstructing the interior of an object from its projections has been a long-standing challenge in CT. Traditional reconstruction
techniques, such as filtered back projection (FBP) \cite{hounsfield1973computerized} and algebraic
reconstruction techniques (ART) \cite{gordon1970algebraic}, often perform poorly under angle constraints due to missing information. FBP is an analytic solution for full-angle CT scans but leaves significant artifacts when working with limited angles. Current solutions to the limited-angle problem encompass a wide range of approaches, such as iterative and model-based solutions. This section will focus on iterative, deep learning, and hybrid methods.

\subsection{Iterative methods}
Iterative methods have long been a cornerstone in addressing limited-angle CT challenges. These approaches, such as ART and simultaneous iterative reconstruction technique (SIRT) \cite{gilbert1972iterative}, reconstruct the image by iteratively minimizing the difference between measured and calculated projections. Since the limited-angle problem is ill-posed, regularization strategies such as total variation (TV) \cite{estrela2016total} and sparsity constraints \cite{candes2006robust} are often used to incorporate prior information, such as piecewise constant solutions or specific shapes. Increasingly, Deep Image Priors (DIP) \cite{ulyanov2018deep} have been used to regularize the reconstructions to favor solutions with more "natural" image statistics \cite{ferreira2023deep,barutcu2021limited}.
These methods are computationally intensive but offer flexibility in integrating priors and physical models of the imaging system.

\subsection{Deep learning methods}
Machine learning has emerged as a tool for tackling limited-angle CT \cite{arndt2023model,germer2023limited,huang2020limited,zhang2016image}.
Neural networks have been used to reconstruct images from limited projection data by training on large datasets of paired sinograms and ground-truth images. However, the scarcity of real-world data in limited-angle scenarios poses a significant challenge for supervised learning.
Sometimes creating synthetic data is possible, which allows researchers to create an arbitrary number of samples with known ground truth images enabling improved supervised models \cite{germer2023limited,wang2020deep, zhou2019limited}.
However, the synthetic data should accurately represent the true images, the imaging system, and noise to generalize to real data. Arguably, creating a synthetic dataset contains as many assumptions as regularization methods in iterative approaches.

\subsection{Hybrid approaches}
Hybrid approaches combine iterative methods and machine learning. These methods aim to leverage the strengths of both approaches to improve reconstruction quality. For instance, iterative methods can provide a good initial estimate, while machine learning techniques can refine this estimate by removing artifacts or enhancing details.

One common hybrid approach is to use one reconstruction algorithm, such as FBP, ART, or SIRT, to create the first reconstruction, and then use a neural network, often a U-net, to remove artifacts or noise from the initial reconstruction \cite{huang2020limited,xu2024hybrid,wang2020deep,arndt2023model}. The benefit of this approach compared to learning the end-to-end reconstruction is that the initial reconstruction can provide a good starting point for the neural network, the task is simplified, and the input space is fixed to the image size rather than varying with the sinogram size.

Additionally, some hybrid approaches use machine learning to learn regularization terms or hyperparameters that are then incorporated into the iterative reconstruction process \cite{kim2019extreme,shen2018intelligent}. For example, a neural network can be trained to estimate the best total variation regularization strength based on the input sinogram, allowing for adaptive regularization that improves reconstruction quality.

\section{Methodology}
\subsection{Problem}
In the limited-angle CT problem, we are provided with projection
measurements of an object from a restricted range of angles (sinogram),
and our goal is to reconstruct an object's interior.

Formally, let $Y$ be the true (unknown) image, $S$ the measured sinogram, and $R(Y,\bar{\theta})$ the Radon transform operator with measurement angles $\bar{\theta}$. We aim to solve for $Y$ in:
$$S = R(Y,\bar{\theta}).$$

To tackle the issue of ill-posedness, we introduce regularization methods
that incorporate prior knowledge about the image, such as smoothness or
sparsity. Regularization methods help stabilize the reconstruction
process and balance between desired properties. We want to find an image $\hat{Y}$ that produces a sinogram
$\hat{S}$ that is close to the measured sinogram $S$ while also satisfying
the regularization constraints. This can be formulated as:
\begin{equation}
\label{eq:minimization}
\hat{Y} = \argmin_{\bar{Y}}\  \left[ \|R(\bar{Y},\bar{\theta}) - S\|_1 + G(\bar{Y}) \right],
\end{equation}
where $G(\bar{Y})$ is a regularization term that penalizes solutions that do not conform to the prior, and $\| \cdot \|_1$ is the $L1$ norm.

\subsection{Overview}
We propose a gradient-based reconstruction pipeline that uses a combination of regularization methods, such as deep image prior, total variation, sinogram filtering, and patch similarity regularization. The regularization methods are discussed in Section \ref{sec:regularization_methods}.
We optimize the pixel values of the reconstruction, or the weights of a neural network, to minimize the difference between the true sinogram, and the sinogram from our reconstruction.
The pseudo-code is shown in Algorithm ~\ref{alg:reconstruction}.

The algorithm reconstructs an image \(\hat{Y}\) from a sinogram \(S\) and measurement angles \(\bar{\theta}\). It iteratively refines the reconstruction using a neural network \(\mathcal{N}\) with weights \(W\). In each iteration $i$, the network reconstructs an image (\(\hat{Y} \gets \mathcal{N}_{W_i}(S)\)), computes the filtered sinogram of the reconstruction using the Radon transform $R$ and a filter $F$: (\(\hat{S}_f \gets F\bigl(R(\hat{Y},\bar{\theta})\bigr)\)),
and compares $\hat{S}_f$ to the input sinogram's filtered version $S_f \gets F(S)$. The loss function $L(S_f, \hat{S}_f, \hat{Y})$ is then computed, and the network weights are updated using the Adam~\cite{kingma2014adam} optimizer. The algorithm stops after a fixed number of iterations, \(N_{\text{iter}}\), because in real-world scenarios the true image is unavailable, making early stopping conditions unreliable. Common ad hoc methods, such as those based on reconstruction variations \cite{wang2021early}, are thus avoided.

The loss function is defined as:
\begin{equation}
\label{eq:loss_fun}
L(S_f,\hat{S_f}, \hat{Y}) = \|S_f - \hat{S_f}\|_1 + G(\hat{Y}),
\end{equation}
where \(S_f\) is the filtered sinogram, \(\hat{S}_f\) is the filtered sinogram from the reconstruction, \(\hat{Y}\) is the reconstructed image, \(G(\hat{Y})\) is a regularization term detailed in Section \ref{sec:regularization_methods}, and \(\| \cdot \|_1\) is the \(L1\) norm.

\begin{algorithm}
       \label{alg:reconstruction}
       \caption{Reconstruction algorithm}
       \begin{algorithmic}[1]
       \Require Sinogram \(S\), measurement angles \(\bar{\theta}\), initial neural network weights \(W_0\), number of iterations \(N_{\text{iter}}\)
       \Ensure Reconstructed image \(\hat{Y}\)
       
       \State \(S_f \gets \text{F}(S)\) \Comment{Filter the sinogram}
       \State \(i \gets 0\)
       \Repeat
           \State \(\hat{Y} \gets \mathcal{N}_{W_i}(S)\) \Comment{Reconstruct image using neural network}
           \State \(\hat{S}_f \gets F\bigl(R(\hat{Y},\bar{\theta})\bigr)\) \Comment{Compute filtered sinogram from reconstruction}
           \State \(L \gets \text{L}(S_f, \hat{S}_f, \hat{Y})\) \Comment{Compute loss as in Equation~\ref{eq:loss_fun}}
           \State \(W_{i+1} \gets \text{Adam}(W_i, \nabla L)\) \Comment{Update weights using the Adam optimizer}
           \State \(i \gets i+1\)
       \Until{\(i \geq N_{\text{iter}}\)}
       \State \Return \(\hat{Y}\)
       \end{algorithmic}
\end{algorithm}

\iffalse
\begin{figure}
       \centering
       \includegraphics[width=1\textwidth]{img/pipeline.png}
       \caption{An illustration of the proposed reconstruction pipeline. The filtering operations are hidden for clarity.
       \textbf{1.} As input, we are given a sinogram $S$ and the measuring angles $\bar{\theta}$.
       \textbf{2.} We have a neural network with weights $W$ that takes in the raw sinogram and creates an image $\hat{Y}$.
       \textbf{3.} We compute the sinogram from our reconstruction: $\hat{S}=\hat{R}(\hat{Y},\bar{\theta})$.
       \textbf{4.} We adjust the weights $W$ of the neural network to minimize the difference between the filtered input sinogram $S_f$ and the filtered sinogram from our reconstruction $\hat{S_f}$.}
       \label{fig:pipeline}
\end{figure}
\fi

Assuming that the real-world measurement operation $\hat{R}$ is close enough to the ideal Radon transform $R$, our underlying assumption is that if the Radon transform of our reconstruction $\hat{Y}$ is close to the measured sinogram $S \gets \hat{R}(Y,\bar{\theta})$, then the reconstruction $\hat{Y}$ is also close to the true image $Y$: $$R(\hat{Y}, \bar{\theta}) \approx \hat{R}(Y,\bar{\theta}) \implies \hat{Y} \approx Y.$$

The HTC'22 challenge also provides an image of a uniform disk with no holes. We use this disk as a mask to constrain the reconstructions to a disk area since we know the exterior is background. The test images are not necessarily centered and simply using the mask as-is might leave edges outside the mask making them impossible to reconstruct. To account for potential misalignment between the mask and the reconstructed image, we introduce two offset parameters that represent the displacement of the disk's center from the image's center. These offset parameters are optimized alongside the reconstruction, allowing the mask to shift to a better position.
Instead of finding the best discrete mask position, we optimize a continuous offset to the mask's center by using bilinear interpolation to shift the mask, making gradient-based optimization possible.

\subsection{Proposed regularization}
\label{sec:regularization_methods}

We propose a combination of regularization methods to improve the reconstruction quality. We divide the regularization methods into two categories: embedded and explicit regularization. Embedded regularization methods are integrated directly into the reconstruction pipeline, while explicit regularization methods are explicitly modeled inside the regularization term $G(\hat{Y})$ in Equation~\ref{eq:minimization}. The regularization terms are combined as:
$$G(\hat{Y}) = \lambda_{\text{TV}} \cdot \text{TV}(\hat{Y}) + \lambda_{\text{PSR}} \cdot \text{PSR}(\hat{Y})$$
where $\lambda_{\text{TV}}$ and $\lambda_{\text{PSR}}$ are hyperparameters that control the strength of the regularization terms.

\subsubsection{Sinogram filtering}
%\noindent\textbf{Sinogram Filtering}\newline\noindent
\label{sec:sinogram_filtering}
Filtering the sinogram before solving the reconstruction problem
is effective for noise suppression. Rather than directly minimizing
$\|R(\hat{Y},\bar{\theta}) - S\|_1$, we filter the sinogram with a customized ramp filter modulated by a squared sinc function \cite{csheaff_filt_back_proj},
and minimize
\[
\|F\big(R(\hat{Y}, \bar{\theta})\big) - F(S)\|_1,
\]
where \(F\) represents the filtering operation.

\begin{figure}
       \centering
       \begin{minipage}{0.32\textwidth}
              \centering
              \includegraphics[width=\linewidth]{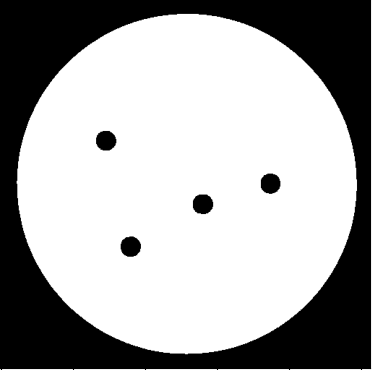}
       \end{minipage}\hfill
       \begin{minipage}{0.32\textwidth}
              \centering
              \includegraphics[width=\linewidth]{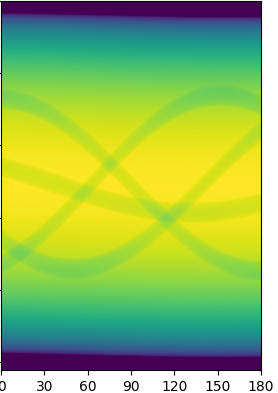}
       \end{minipage}\hfill
       \begin{minipage}{0.32\textwidth}
              \centering
              \includegraphics[width=\linewidth]{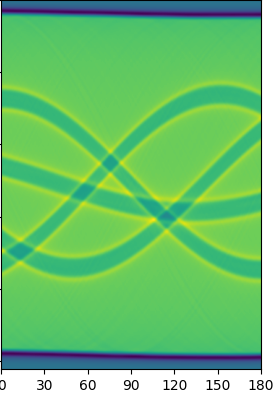}
       \end{minipage}
       \caption{The image shows the true image (left), the sinogram (middle), and the filtered sinogram ($\alpha=5.0$) (right). The filtering reduces high-frequency content which makes the holes more prominent in the sinogram, and mitigates overfitting to noise.}
       \label{fig:demo_sinograms}
\end{figure}

The filter is parameterized by a single parameter $\alpha$, which
controls the amount of filtering.
A smaller $\alpha$ approaches a pure ramp filter, while larger $\alpha$
progressively reduces higher-frequency content. An example of the effect of filtering is shown in Figure~\ref{fig:demo_sinograms}.

Mathematically, the filter is:
\[
r_{\alpha}(\omega) = \left| \frac{2}{\alpha} \sin\left(\frac{\alpha\omega}{2}\right) \right|
\left[ \frac{\sin\left(\frac{\alpha\omega}{2}\right)}{\frac{\alpha\omega}{2}} \right]^2,
\]
where \(\omega\) is the computed by taking the 1D Fourier transform of each projection. After filtering, we take the inverse Fourier transform to obtain the filtered sinogram.
Thus, the sinogram filtering operation is:
\[
F_{\alpha}(S) = \mathcal{F}^{-1}\left[ r_{\alpha}(\mathcal{F}(S)) \right],
\]
where \(\mathcal{F}\) is the Fourier transform, and \(\mathcal{F}^{-1}\) is the inverse Fourier transform.

\noindent
\subsubsection{Deep Image Prior}
% \textbf{Deep Image Prior}\newline\noindent
Deep Image Prior (DIP)~\cite{ulyanov2018deep}
leverages the structure of convolutional neural networks as
an implicit prior, favoring images with more natural statistics.
Instead of directly optimizing the pixel values of $\hat{Y}$, we instead optimize the neural network parameters so that 
$$
\hat{Y} = \mathcal{N}_W(S) \quad \text{with} \quad W = \argmin_{\bar{W}}  \bigl[ \left\lVert F\bigl(R(\mathcal{N}_{\bar{W}}(S))\bigr) - F(S) \right\rVert_1 + G\bigl(\mathcal{N}_{\bar{W}}(S)\bigr) \bigr].
$$
where $\mathcal{N}_{W}$ is a convolutional neural network (CNN) with weights $W$, $S$ is the input sinogram, $\hat{Y}$ is the reconstructed image, $F$ is the sinogram filtering operator, and $G$ is the regularization term.

By constraining the image to be a CNN output, we bias the optimization process towards solutions with natural image statistics and spatial relationships, rather than overfitting to noise in the sinogram. DIP has been successfully applied to various imaging
tasks~\cite{baguer2020computed,barutcu2021limited}.

The backbone of our proposed DIP is based on the ConvNeXt architecture, following the
winners of HTC'22~\cite{germer2023limited}.
However, we avoid padding the sinogram and adding another channel to denote where the sinogram is since our use case is different and the DIP only needs to work for a single-size sinogram, thus making the input space much smaller.
Using grid-search, we individually adapt the convolution parameters, such as padding and kernel size for any given sinogram.

\subsubsection{Total variation}
% \textbf{Total Variation}\newline\noindent
Total variation (TV) regularization is widely
used in CT reconstruction. It encourages piecewise constant
solutions by minimizing the $L1$ norm of the image gradient \cite{estrela2016total}.
We calculate the total variation of an image $\hat{Y}$ using the formula:

\begin{equation}
TV(\hat{Y}) = \frac{\sum_{i,j} \left| (\nabla \hat{Y})_{x_{i,j}} \right| + \left| (\nabla \hat{Y})_{y_{i,j}} \right|}{N},
\end{equation}
where \((\nabla \hat{Y})_{x_{i,j}}\) and \((\nabla \hat{Y})_{y_{i,j}}\) are the horizontal and vertical gradients of the image \(\hat{Y}\) at pixel location \((i, j)\), respectively, and \(N\) is the number of pixels in the image.

%\textbf{Patch similarity}\newline\noindent
\subsubsection{Patch Similarity}
To improve our reconstructions, we add patch similarity regularization (PSR), a patch-based regularization step that serves as an extra structural prior. We begin by training an autoencoder (AE) on image patches that match the target domain (for example, shapes similar to those in the HTC'22 challenge). This autoencoder learns a compact, low-dimensional representation of what valid patches should look like.

When optimizing, we partition the current reconstruction \(\hat{Y}\) into patches and run each one through the autoencoder. We then measure the difference between the original patches and their autoencoded reconstructions. This difference is used as a penalty that nudges the overall reconstruction toward having patches that resemble the training data.

\begin{algorithm}
       \caption{PSR penalizes patches that do not resemble the training data.}
       \label{alg:patch_sim}
       \begin{algorithmic}[1]
       \Require Current solution $ \hat{Y} \in \mathbb{R}^{n \times n}$, patch size $p \times p$, stride $s = p$
       \Function{PSR}{$\hat{Y}, p, s$}
           \State $P \gets \textsc{SplitIntoPatches}(\hat{Y}, p, s)$ \Comment{Divide $\hat{Y}$ into patches}
           \State $P_r \gets \textsc{EncodeDecodePatches}(P)$ \Comment{Encode and decode patches}
           \State $penalty \gets \textsc{Mean}\Bigl(|P - P_r|\Bigr)$ \Comment{Compute the mean absolute difference}
           \State \Return $penalty$
       \EndFunction
       \end{algorithmic}
\end{algorithm}

\begin{figure}[t]
       \centering
       \includegraphics[width=\textwidth]{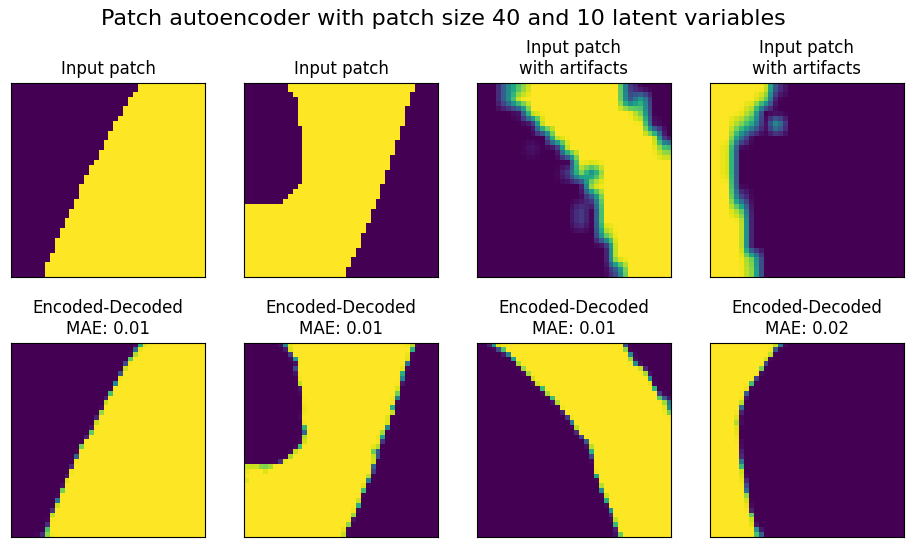}
       \caption{Examples of patches encoded and decoded by the autoencoder. MAE is the mean absolute error between the true and reconstructed patches.
       Top row: Input patches. Bottom row: Reconstructed patches.
       The autoencoder learns to encode and decode patches that are similar to its training data. If the autoencoder is restrictive enough, it cannot encode noise or artifacts, thus providing a strong prior for the reconstruction.}
       \label{fig:patch_encoder_examples}
\end{figure}

The autoencoder is built to capture the essential features of a patch while filtering out unusual artifacts. Figure~\ref{fig:patch_encoder_examples} shows examples of patches and their reconstructions by the autoencoder. As can be seen, the autoencoder leaves realistic patches unchanged and removes noise and artifacts from unrealistic patches.

For training, we create eight images that mimic the HTC'22 phantoms (using the approach described in \cite{germer2023limited}) and split them into overlapping patches with stride $\lfloor \tfrac{\text{patch\_size}}{5} \rfloor$.
We then train the autoencoder to minimize the binary cross-entropy \cite{shannon1948mathematical} between the original and reconstructed patches. The autoencoder is trained with a batch size of 32, using the Adam optimizer with a learning rate of 0.001, for 100 epochs.
The autoencoder architecture is a symmetric encoder-decoder model with three convolutional layers. We keep the latent dimension at \(\lfloor \tfrac{\text{patch\_size}}{4} \rfloor\), which we found small enough to force the autoencoder to learn a compact representation of the patches. 
% We train four different autoencoders using the same architecture and training settings, shown in Table~\ref{tab:patch_data_stats}.

% \begin{table}
% \centering
% \caption{A smaller patch size gives simpler patches, but more of them.}
% \label{tab:patch_data_stats}
% \begin{tabular}{lcccc}
% \toprule
% Patch size & Stride & Latent variables & Patches per image & Total patches \\
% \midrule
% 15x15 & 3 & 3 & 27561 & 220488 \\
% 20x20 & 4 & 5 & 15376 & 123008 \\
% 30x30 & 6 & 7 & 6561 & 52488 \\
% 40x40 & 8 & 10 & 3600 & 28800 \\
% \bottomrule
% \end{tabular}
% \end{table}

\section{Experiments}

Our solution is implemented in Python using PyTorch, with TorchRadon \cite{torch_radon}
for differentiable Radon transforms. The entire pipeline is available in an open-source
repository at: https://github.com/ilmari99/data-efficient-lact.

\subsection{Data}
\label{sec:data}
\begin{figure}
       \centering
       \includegraphics[width=1.0\textwidth]{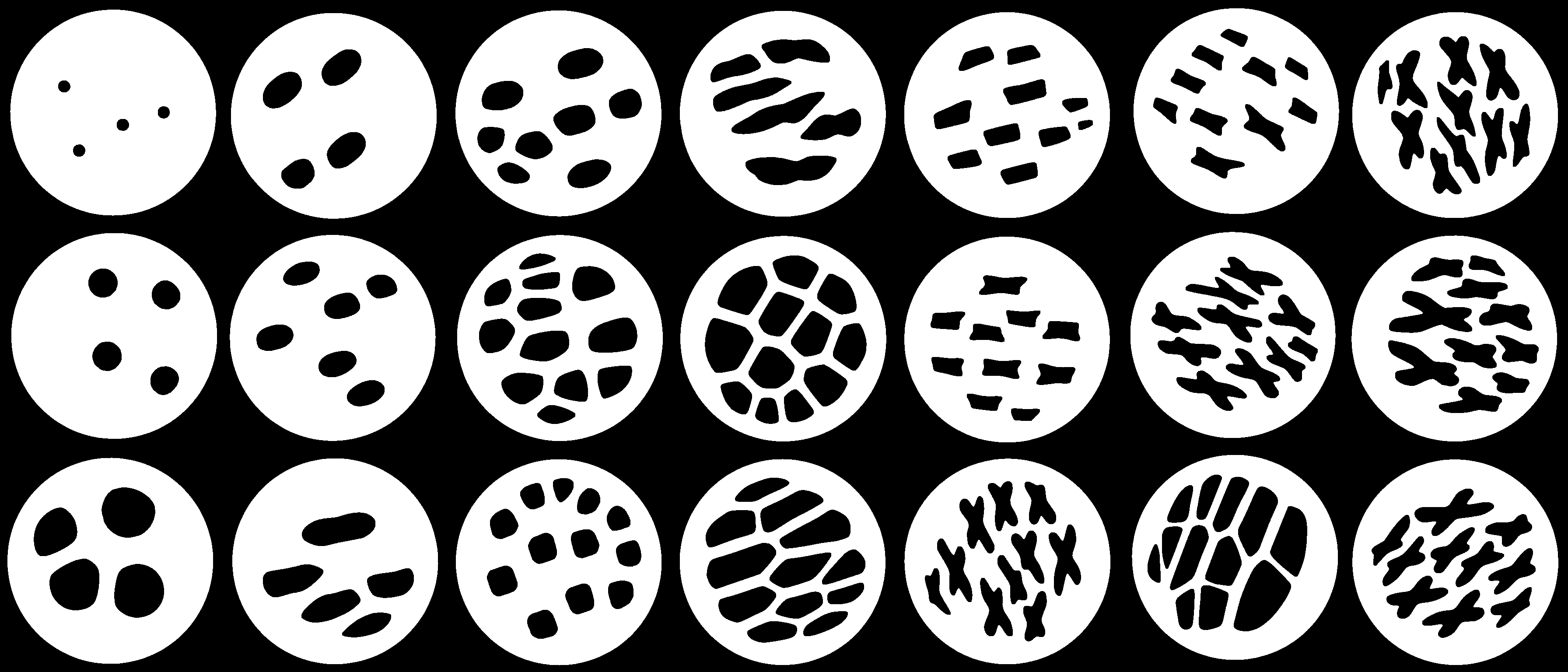}
       \caption{All 21 phantoms used in the HTC'22 challenge.
       From left to right: Level 1-7. From top to bottom: Phantom a, b, c.}
       \label{fig:all_phantoms}
\end{figure}

The challenge's test dataset has seven levels, each reducing the scan angle by
$10^\circ$, starting from $100^\circ$ (level 1) down to $30^\circ$ (level 7).
Three phantoms (a, b, c) were provided for each level.
Figure~\ref{fig:all_phantoms} gives an overview of all 21 phantoms.

The challenge provided four training/demo images and their
full sinograms to help tune algorithms. In our experiments we use these demo images with 30-degree sinograms to perform a hyperparameter search.

The evaluation metric is to find a solution $\hat{Y}$
that maximizes the Matthews correlation coefficient (MCC) \cite{matthews1975comparison},
also known as the $\phi$-coefficient,
between the true image $Y$ and the reconstructed image $\hat{Y}$.

\subsection{Tests and validation}

We conduct a hyperparameter search using the four training/demo images. To ensure an unbiased and reproducible search of the hyperparameter space, we use random sampling from a heuristically defined search space. Based on the search results, we perform an ablation study to assess the contribution of each regularization technique. For each regularization method, we identify the best-performing hyperparameter set from trials where that specific regularization was excluded.

We then compare the performance of these ablated configurations against the full model with all regularization methods included. The three best-performing hyperparameter sets are evaluated on the HTC'22 test dataset. Finally, we compare our results against the official challenge results and common baseline methods.
We validate our test environment by comparing the HTC'22 winner's official results, to what our environment reported for the winner's model. We find that our environment is able to reproduce the winner's results within floating-point error, giving us confidence in our setup.

\section{Results}

\subsection{Hyperparameter and ablation study}
We study the importance of each regularization method by finding the best trials, where one regularization method is excluded. We then compare the MCC scores of these ablated trials against the full model that uses TV, sinogram filtering, DIP, and PSR.

The bigger the difference from the full model, the more important the regularization technique is. From the table, we can see that DIP has the biggest impact, followed by filtering, PSR, and TV.
The results of the ablation study are shown in Table~\ref{tab:ablation_study}.

\begin{table}[t]
\centering
\caption{Ablation study results. The best hyperparameter set for each ablated configuration is used. In each row, the ablated hyperparameters are in \textcolor{blue}{blue}.}
\label{tab:ablation_study}
\begin{tabular}{lcccccccc}
\toprule
Name & DIP & $F_\alpha$ & $\lambda_{TV}$ & $\lambda_{PSR}$ & Patch size & lr & $N_{iter}$ & MCC \\
\midrule
No DIP & \textcolor{blue}{False} & 5.5 & 0.1 & 0.1 & 30 & 0.2 & 300 & 3.07 \\
No Filter & True & \textcolor{blue}{0.0} & 0.5 & 0.1 & 30 & 0.01 & 1200 & 3.23 \\
No TV & True & 5.5 & \textcolor{blue}{0.0} & 0.2 & 20 & 0.001 & 1200 & 3.44 \\
No PSR & True & 5.0 & 0.5 & \textcolor{blue}{0.0} & \textcolor{red}{-} & 0.001 & 1200 & 3.44 \\
\midrule
\cellcolor{mistyrose}{Full} & \cellcolor{mistyrose}{True} & \cellcolor{mistyrose}{6.0} & \cellcolor{mistyrose}{0.01} & \cellcolor{mistyrose}{0.2} & \cellcolor{mistyrose}{40} & \cellcolor{mistyrose}{0.001} & \cellcolor{mistyrose}{400} & \cellcolor{mistyrose}{\textbf{3.55}} \\
\bottomrule
\end{tabular}
\end{table}

The HTC'22 competition allowed teams to submit as many solutions as their team had members. Thus, we also choose three of the best hyperparameter sets from our hyperparameter search and evaluate them on the test dataset. The best-performing of these three hyperparameter sets is then compared to the baseline methods and the HTC'22 results, and is referred to as "Our method" in the results.

\subsection{Quantitative results}
Table~\ref{tab:mcc_results} shows the total MCC score achieved on each level of the challenge with different methods. Our method outperforms the average HTC'22 competitors on all levels and is competitive with the best challenge results.
Though our method is not better than the best challenge result, it \textbf{uses only 0.006\% of the data compared to the HTC'22 winner} who use 200,000 synthetic images. This makes our method a strong contender for real-world applications where data is scarce. In Table~\ref{tab:mcc_results} "FBP" refers to filtered back projection with Otsu thresholding~\cite{yousefi2011image} and using a mask to constrain the reconstruction to the disk area, "No reg." refers to our method without any regularization, "Avg. in HTC'22" and "Best in HTC'22" refer to the average and best results in the HTC'22 challenge, and "Our method" refers to the best hyperparameter set in the top-3 trials.

\begin{table}[t]
       \centering
       \caption{MCC scores of our method compared to baseline and challenge results.}
       \label{tab:mcc_results}
       \begin{tabular}{lccccccc}
       \toprule
       Method & Level 1 & Level 2 & Level 3 & Level 4 & Level 5 & Level 6 & Level 7 \\
       \midrule
       FBP & 2.79 & 2.75 & 2.52 & 2.38 & 2.44 & 1.82 & 1.69 \\
       No reg. & 2.68 & 2.77 & 2.66 & 2.20 & 2.50 & 1.53 & 1.59 \\
       Avg. in HTC'22 & 2.79 & 2.69 & 2.62 & 2.57 & 2.61 & 2.28 & 2.07 \\
       Best in HTC'22 & 2.96 & 2.97 & 2.93 & 2.92 & 2.93 & 2.81 & 2.41 \\
       \midrule
       \cellcolor{mistyrose}{Our method} & \cellcolor{mistyrose}{2.90} & \cellcolor{mistyrose}{2.88} & \cellcolor{mistyrose}{2.78} & \cellcolor{mistyrose}{2.54} & \cellcolor{mistyrose}{2.75} & \cellcolor{mistyrose}{2.53} & \cellcolor{mistyrose}{2.17} \\
       \bottomrule
       \end{tabular}
\end{table}

\begin{figure}[H]
       \centering
       \includegraphics[width=0.75\textwidth]{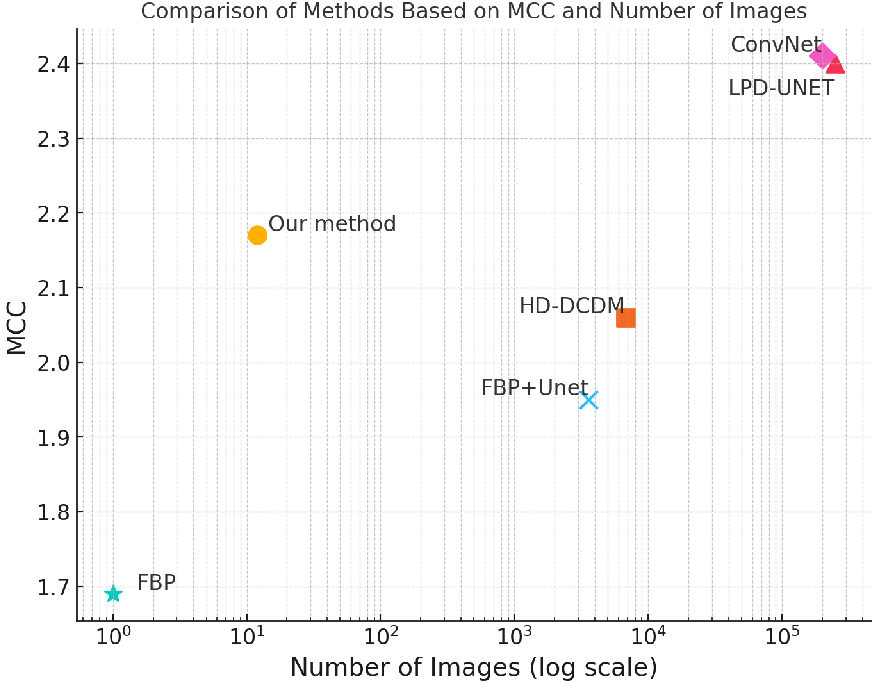}
       \caption{Accounting for the amount of data used, our method compares favorably to HTC'22 submissions. The x-axis shows the number of data points used in training, and the y-axis shows the MCC score on the level 7 test set.}
       \label{fig:performance_comp}
\end{figure}

Figure \ref{fig:performance_comp} shows how our method compares to the HTC'22 submissions and FBP when accounting for the amount of data used. The x-axis shows the number of data points used in training, and the y-axis shows the MCC score on the level 7 test set. The methods we compared to are FBP, HD-DCDM \cite{wang2023hd}, ConvNet \cite{germer2023limited}, LPD-UNET \cite{arndt2023model}, and FBP+UNET \cite{wang2023deep}.
Our method is competitive with the HTC'22 submissions, despite using only 12 data points, compared to the HTC'22 winner's 200 000 synthetic images.

\subsection{Qualitative results}
\begin{figure}[H]
\centering
\includegraphics[width=0.8\textwidth]{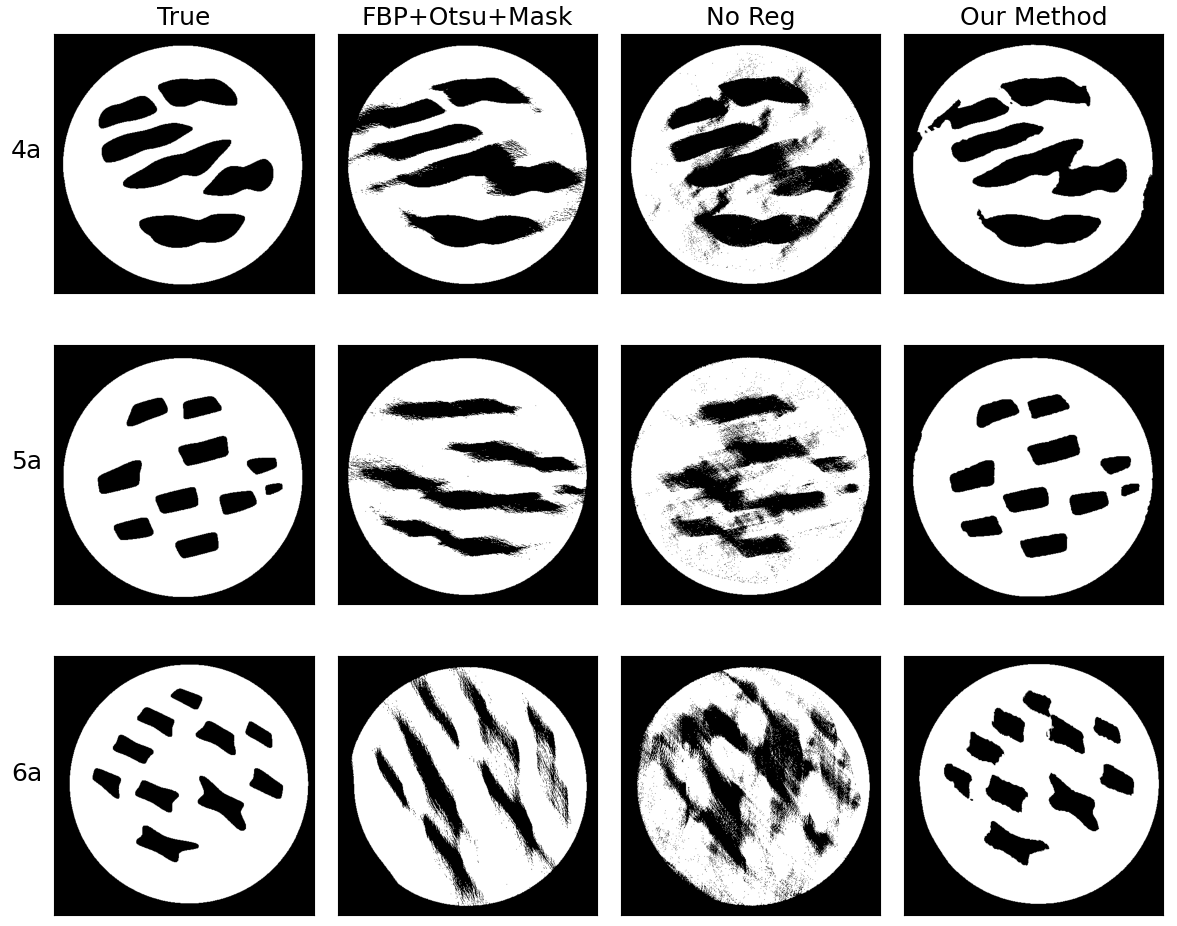}
\caption{Reconstruction images of the binary disk from the HTC'22 challenge. From left to right: True image, FBP + Otsu + Mask, our method with no regularization, and our full method.}
\label{fig:qualitative_results}
\end{figure}

Figure~\ref{fig:qualitative_results} shows the qualitative results of our method compared to the baseline methods. As visible in the figure, FBP and our method with no regularization struggle with artifacts and noise, while our method produces clean and more accurate reconstruction.

\subsection{Discussion}

Our results demonstrate that high-quality CT reconstructions can be achieved with minimal data by using optimization-based reconstruction with good regularization. Despite using only 12 data points, our approach competes with state-of-the-art deep learning models trained on 200,000 synthetic images, making it highly applicable in data-scarce scenarios. Ablation studies highlight the importance of Deep Image Prior, sinogram filtering, PSR, and total variation in optimization. While our method is more computationally intensive than pre-trained deep learning models, taking approximately 30 seconds on an NVIDIA RTX 3060, it eliminates the need for large labeled datasets and generalizes better to real-world imaging conditions.
Future work could focus on testing our method on datasets with more complicated structures and finding optimal stopping criteria for the reconstruction.

\bibliographystyle{splncs04}
\bibliography{refs.bib}

\end{document}